\documentclass[letterpaper, 10 pt, conference]{ieeeconf}  

\IEEEoverridecommandlockouts                              

\usepackage{graphicx} 
\usepackage{amsmath} 

\title{\LARGE \bf
Learning Highly Dynamic Behaviors for Quadrupedal Robots
}

\author{Chong Zhang$^{1}$, Jiapeng Sheng$^{1}$, Tingguang Li$^{1}$, He Zhang$^{1}$, \\ Cheng Zhou$^{1}$, Qingxu Zhu$^{1}$, Rui Zhao$^{1}$, Yizheng Zhang$^{1}$, Lei Han$^{1}$
\thanks{$^{1}$Chong Zhang, Jiapeng Sheng, Tingguang Li, He Zhang, Cheng Zhou, Qingxu Zhu, Rui Zhao, Yizheng Zhang and Lei Han are with Tencent Robotics X Laboratory,
        Shenzhen, Guangdong, China.
        {\tt\small chongzzhang@tencent.com}}
}

\begin{document}

\maketitle
\thispagestyle{empty}
\pagestyle{empty}

\begin{abstract}

Learning highly dynamic behaviors for robots has been a longstanding challenge. Traditional approaches have demonstrated robust locomotion, but the exhibited behaviors lack diversity and agility. They employ approximate models, which lead to compromises in performance. Data-driven approaches have been shown to reproduce agile behaviors of animals, but typically have not been able to learn highly dynamic behaviors. In this paper, we propose a learning-based approach to enable robots to learn highly dynamic behaviors from animal motion data. The learned controller is deployed on a quadrupedal robot and the results show that the controller is able to reproduce highly dynamic behaviors including sprinting, jumping and sharp turning. Various behaviors can be activated through human interaction using a stick with markers attached to it. Based on the motion pattern of the stick, the robot exhibits walking, running, sitting and jumping, much like the way humans interact with a pet.

\end{abstract}

\section{INTRODUCTION}

Animals exhibit remarkable agility in the wild. In order to escape from predators, they perform various behaviors such as sprinting, jumping and high-speed sharp turning. Enabling robots to reproduce these behaviors has been a long-standing challenge. As robots move at high speed, several physical considerations begin to influence the dynamics of the robot, including the enforcement of motor limit, the regulation of contact force and posture balance during flight phases~\cite{margolis2022rapid}.

Manually designing controllers to resolve these issues typically requires a lengthy design process and tedious parameter tuning~\cite{bellicoso2017dynamic,chi2022linearization,zhou2023max}. Conventional approaches break the system into manageable submodules and each submodule generates reference values for the next~\cite{kolter2008control,kalakrishnan2010fast}. Animal data is utilized in~\cite{kang2021animal,kang2022animal} to allow the robot to reproduce relatively agile behaviors. However, these methods employ approximate models for each module and lead to compromises in performance. Controllers based on trajectory optimization suffer from expensive computational costs due to the optimization needs to be performed at execution time~\cite{gehring2016practice,neunert2017trajectory,carius2018trajectory,winkler2018gait}.

Learning-based approaches, such as Reinforcement Learning (RL), promise the potential to overcome the limitations~\cite{peng2018deepmimic,escontrela2022adversarial,ji2022hierarchical,smith2023learning}. The work in~\cite{margolis2022rapid} has shown a neural network controller can push a small quadruped to the limits of its agility, achieving high-speed mobility. However, the task reward function they used is linear and angular velocity tracking, thus the learned motion has limited diversity and is not guaranteed to exhibit animal-like behaviors. In~\cite{peng2020learning}, the authors proposed an imitation learning system that enables the quadruped to learn agile locomotion skills by imitating real animals. However, their model can only track low-speed reference trajectories and the robot cannot be guided in a controllable way. Primitive skill reuse has been first investigated in character animation~\cite{levine2012continuous,ling2020character,wagener2022mocapact} and then expanded to robotics~\cite{huang2022creating,bohez2022imitate}. These approaches compress the primitive motion to a continuous latent space, and navigate this space to accomplish use-specific tasks. In~\cite{bohez2022imitate}, the researchers investigated the use of prior knowledge of animal behaviors to learn reusable locomotive skills in quadruped robots. However, the behaviors performed by their robot also lacks agility and diversity. In~\cite{li2023learning}, the researchers proposed a learning-based approach to learn traversing challenging terrains by imitating animals. The robot can perform natural-looking behaviors on different terrains like stairs and slopes, but in relatively low-speed mode.

In this paper, we propose a data-driven approach that enables the robot to perform various highly dynamic behaviors robustly and human can interact with the robot by actively triggering these skills.
Our model is trained in simulation and deployed on a quadrupedal platform designed by our team~\cite{chi2022linearization,zhou2023max}.
The main contribution of our paper is as follows. First, we propose a learning framework that allows the robot to imitate a versatile of motions collected from real animals. The framework utilizes a vector quantized controller~\cite{han2023lifelike,zhu2023neural} and could robustly track high-speed motion data including sprinting, jumping and sharp turning. Second, various behaviors can be activated, allowing humans to interact with the robot in a controllable manner. Humans can utilize a stick with markers attached to it to interact with the robot, much like the
way humans interact with a pet. Third, several generic techniques have been proposed to improve the tracking accuracy and robustness of the robot. The tracking accuracy of high-speed motion is improved by utilizing prioritized sampling and early termination. Prioritized sampling is enabled to encourage the model to exploit more highly dynamic motions during training. Early termination provides another means of discouraging undesired behaviors. The robustness of our model deployed on the real robot is achieved by modifying the action space commonly used in previous work~\cite{peng2018deepmimic,tan2018sim}. The action space is represented as the change of joint angular velocities in our model, compared with previous approaches the new representation achieves more fluid and robust motion.

\section{METHODOLOGY}

\begin{figure} [t]
    \includegraphics[width=\linewidth]{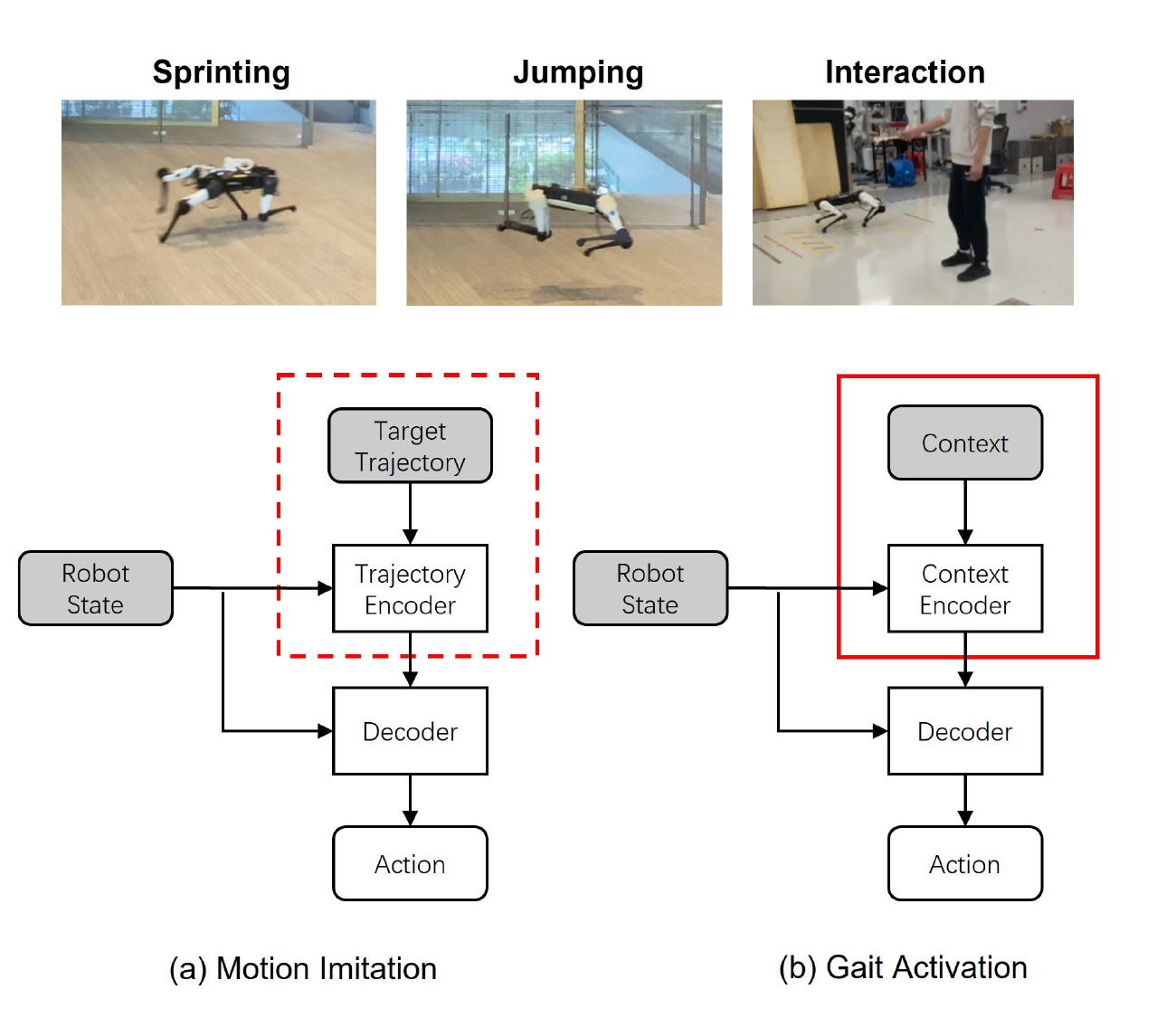}
    \caption{The framework of proposed model. It consists of two stages: (a) motion imitation and (b) gait activation. In motion imitation stage, the model is trying to imitate animal behaviors. In gait activation stage, humans can interact with the robot by triggering various gaits. The module in red dashed rectangle in (a) is replaced by module in red solid rectangle in (b), the remaining module is shared by (a) and (b).}
    \label{fig:figure1}
    \vspace{-0.3cm}
\end{figure}

The general architecture of our model is shown in Fig.~\ref{fig:figure1}. The model is learned in two stages. In the first stage, we use a trajectory encoder and decoder to imitate the animal motion data. The trajectory encoder encodes the target trajectory as a latent control signal, the decoder outputs action based on the current robot state and the latent control signal. In the second stage, we freeze the parameters of the decoder and replace the trajectory encoder by a context encoder, which tries to learn appropriate latent control signals to navigate the robot based on desired commands.

\subsection{Data Acquisition and Processing}

\begin{figure} [t]
    \centering
    \includegraphics[width=0.8\linewidth]{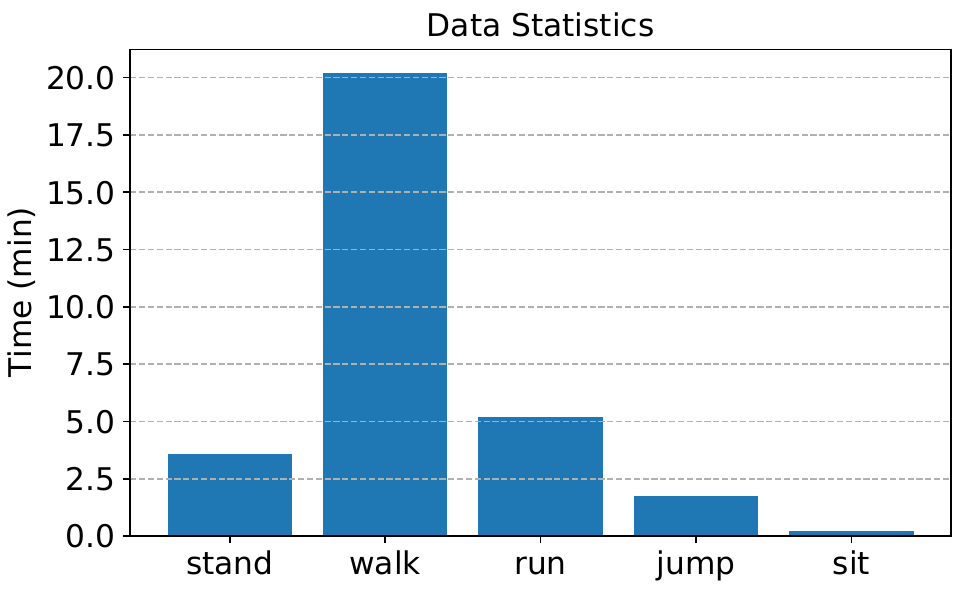}
    \caption{Statistics of collected motion data from a medium-sized Labrador retriever. Horizontal axis represents gait type, vertical axis represents the duration in minutes.}
    \label{fig:figure_data_stat}
    \vspace{-0.5cm}
\end{figure}

The training data is collected using motion capture system. We use a medium-sized Labrador retriever as the subject and capture locomotive data of various gaits. The data are saved according to gait type. The subject is guided by a dog trainer to follow various instructions. The gaits include standing, walking, running jumping and sitting. Each gait is repeated 4-6 times to ensure data diversity. The moving trajectories consist of square, circle and straight line. Upon completion, we have approximately 0.5 hour of motion data. We manually label the gait at frame level and plot the statistics in Fig.~\ref{fig:figure_data_stat}. Horizontal axis represents gait type, vertical axis represents the duration in minutes. Since the morphology of real animal is different from the robot, we retarget the source motion to the robot using inverse kinematics~\cite{gleicher1998retargetting}.

\subsection{Motion Imitation}

The motion imitation module consists of a trajectory encoder, a decoder and a latent embedding space. The encoder and decoder are parameterized using neural networks and the embedding space is a \textit{K}-way categorical discrete space, where \textit{K} is the size of the space. The benefits of using discrete over continuous latent space has been shown to improve exploration efficiency for downstream tasks~\cite{zhu2023neural}. The input to the encoder includes current proprioception of the robot and target trajectories within one second time window in the future. The encoder encodes the current robot state and target future state as a control signal in latent space. The current proprioception of the robot and the latent control signal serve as the input to the decoder. The output of the decoder is processed and fed into a proportional-derivative (PD) controller. The PD controller then computes the joint torque for each actuator. Compared with direct output joint torques, PD controllers have been shown to improve performance and learning speed for motor control tasks~\cite{peng2017learning}.

The proprioception of the robot $\mathbf{s}_t^\text{p}$ at timestep \textit{t} is the concatenation of three consecutive states spanning from timestep \textit{t-2} to \textit{t} and history actions from timestep \textit{t-3} to \textit{t-1}. The state at each timestep consists of joint angles, joint angular velocities, root angular velocities and gravity vector. The gravity vector is used to represents the roll and pitch of the robot~\cite{lee2020learning}. Root angular velocities are expressed under the robot coordinate system. The history action is the previous output from the decoder. The future trajectories $\mathbf{s}_t^\text{f}$ at timestep \textit{t} specifies the target pose of the robot. It consists of states of reference motion located 0.03s, 0.06s, 0.3s and 1s relative to current timestep. Each reference state includes target joint angles, target root position and orientation expressed under the robot coordinate system. The input to the encoder is the concatenation of $\mathbf{s}_t^\text{p}$ and $\mathbf{s}_t^\text{f}$, the output of the encoder $\mathbf{z}_t$ is mapped to the nearest element of the embedding space as $\mathbf{c}_{t,k}$, where $\mathbf{c}_{t,k}$ is the \textit{k}-th element in the embedding space and $k=\arg\min_j||\mathbf{z}_t-\mathbf{c}_{t,j}||_2^2$. The decoder $\pi(\mathbf{a}_t|\mathbf{s}_t^\text{p},\mathbf{c}_{t,k})$ is conditioned on the proprioception and the embedding vector $\mathbf{c}_{t,k}$ and outputs the distribution of action $\mathbf{a}_t$.

We try two types of action space and compare the tracking performance under highly dynamic condition. The first type of action space $\mathbf{a}_t^{\text{I}}$ represents the change of joint angles and is obtained from the output of the decoder. The target joint angles for the PD controller $\mathbf{q}_t^*$ are computed as:
\begin{equation}
    \mathbf{q}_t^* = \mathbf{q}_t + \mathbf{a}_t^{\text{I}}
\end{equation}
where $\mathbf{q}_t$ is the current joint angles of the robot. The target joint angular velocities $\dot{\mathbf{q}}_t^*$ are set to zero~\cite{peng2018deepmimic,tan2018sim}.

The second type of action space ${a}_t^{\text{II}}$ represents the change of joint angular velocities. The target angular joint velocities and the target joint angles are computed as:
\begin{equation}
    \dot{\mathbf{q}}_t^* = \dot{\mathbf{q}}_t + \mathbf{a}_t^{\text{II}}
\end{equation}
\begin{equation}
    \mathbf{q}_t^* = \mathbf{q}_t + \dot{\mathbf{q}}_t^* \Delta t
\end{equation}
where $\Delta t$ is the timestep. The current joint angles, angular velocities of the robot, target joint angles and the target joint angular velocities are fed to the PD controller.

The reward for motion tracking task is defined as:
\begin{equation}
    r_t^\text{tr}=r_t^\text{i} + \alpha r_t^\text{q}
\end{equation}
where $r_t^\text{i}$ is the imitation reward, $r_t^\text{q}$ is the vector quantized reward and $\alpha$ is the weighting coefficient. The imitation reward is the weighted sum of pose reward, velocity reward, end-effector reward, root pose reward and root velocity reward. It is defined the same as in~\cite{peng2020learning}. The vector quantized reward is define as~\cite{van2017neural}:
\begin{equation}
    r_t^\text{q} = -\|\text{sg}[\mathbf{z}_t]-\mathbf{c}_{t,*}\|_2^2 - \beta \|\mathbf{z}_t - \text{sg}[\mathbf{c}_{t,*}]\|_2^2
\end{equation}
where \text{sg} is the stopgradient operator, $\mathbf{c}_{t,*}$ is the selected embedding vector and $\beta$ is a coefficient. The first term tries to move the embedding vectors towards the encoder output, the second term is used to make sure the encoder output commit to the embedding vectors. We use proximal-policy optimization (PPO) to update the model parameters~\cite{schulman2017proximal}.

Imitating highly dynamic behaviors such as jumping is challenging. We use prioritized sampling (PS) and early termination (ET) to facilitate motion imitation. Prioritized sampling is enabled to encourage the model to exploit more data with low tracking accuracy. During training, at the beginning of each episode, a particular motion data is selected according to a performance score. The score is computed as the inverse of normalized cumulative imitation reward. The early termination provides another means of reward shaping to discourage undesired behaviors~\cite{peng2018deepmimic}. We find that setting obstacles under the flight trajectories encourages the learned policy to achieve consistent peak height during flight phase with animal data. During training, the peak heights of the retargeted jumping data are detected and obstacles are placed with proper pose. The shape of the obstacle is a cube with length, width and height equal to 0.5m, 0.02m and 0.2m respectively. The size of the cube is chosen to avoid collision with the robot during jumping. The obstacle position is computed by projecting the animal's position at peak height on the ground and the orientation is set equal to the animal's orientation at peak height. Early termination is triggered when detection of a collision, characterized by the robot's body or legs making contact with the cube.

\begin{figure} [t]
    \centering
    \includegraphics[width=\linewidth]{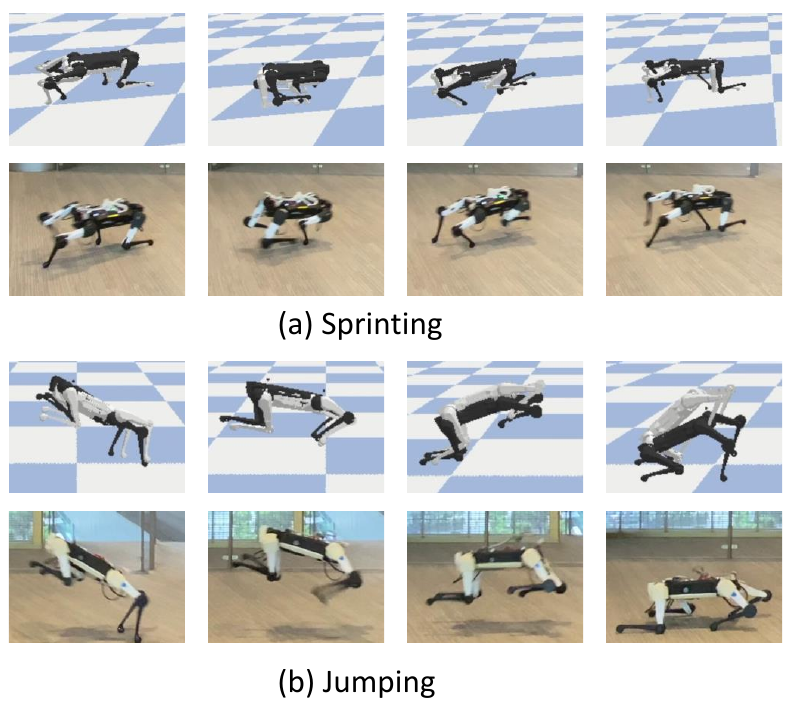}
    \caption{Snapshots of learned robot behaviors. (a) Snapshots of sprinting. (b) Snapshots of jumping}
    \label{fig:figure_animation}
    \vspace{-0.5cm}
\end{figure}

\subsection{Gait Activation}

Gait activation is achieved by removing the trajectory encoder and stacking another context encoder on top of the decoder. The inputs to the context encoder include the current proprioception $\mathbf{s}_t^\text{p}$ and context state $\mathbf{s}_t^\text{c}$. The context state consists of current linear velocity, current root height of the robot, target linear velocity and target height. During training, target direction is uniformly sampled between 0 and $2\pi$, target speed in horizontal plane is uniformly sampled between 0 and 3m/s. Target height is uniformly sampled from three values 0.1m (sit), 0.3m (walk and run) and 0.5m (jump). The linear velocity and root height of the robot are obtained using a motion capture system when deployed on real system. The output of the encoder is a categorical distribution that is used to sample the appropriate embedding vector from the discrete embedding space learned in previous section. The selected embedding $\mathbf{c}_{t,*}$ and $\mathbf{s}_t^\text{p}$ are fed to the decoder.

The context encoder is parameterized with a neural network and is updated using PPO. The reward is defined as:
\begin{equation}
\label{eqn:context reward}
    r_t^\text{at}=r_t^\text{c} + \gamma r_t^\text{g}
\end{equation}
where $r_t^\text{c}$ is the context reward to align the robot liner velocity and root height with the desired velocity and height and is computed similar with~\cite{ling2020character}, $r_t^\text{g}$ is a generative adversarial reward that is used to encourage the produced motion similar to motions learned from animal data~\cite{ho2016generative} and $\gamma$ is the coefficient.
The context reward is defined as:

\begin{equation}
\label{eqn:context reward}
    r_t^\text{c} = e^{\cos(\theta_t - \theta_t^\text{d})-1} \times e^{-|v_t-v_t^\text{d}|} \times e^{-|h_t-h_t^\text{d}|}
\end{equation}
where $\theta_t$ is the current facing direction of the robot projected on horizontal plane, $\theta_t^\text{d}$ is the desired facing direction, $v_t$ is the speed of the robot in horizontal plane, $v_t^\text{d}$ is the desired speed in horizontal plane, $h_t$ is the root height of the robot and $h_t^\text{d}$ is the desired height. The imitation reward is defined as:

\begin{equation}
\label{eqn:context reward}
    r_t^\text{g}=-\log(1-D(\mathbf{s}_t^\text{p},\mathbf{a}_t))
\end{equation}
where $D(\mathbf{s}_t^\text{p},\mathbf{a}_t)$ is a discriminator to classify the state-action pair is produced from gait activation stage or motion imitation stage. We use the same objective function for the discriminator as in~\cite{peng2021amp}. Compared to directly learning gait activation with the imitation reward, our two phase learning framework is able to reuse the the skills learned in the first stage for various downstream tasks.

\subsection{Real World Transfer}

Our model is trained in simulation and deployed on a quadrupedal platform. To facilitate the transfer of learned policy from simulation to real world, we follow most existing approaches proposed by ~\cite{peng2018deepmimic,tan2018sim,hwangbo2019learning}. We randomize the terrain friction, the actuators' torque limit, and periodically apply random forces on the root of the robot. At the beginning of  each episode, the lateral friction of the terrain is uniformly sampled between 0.9 to 2.0, the torque limit of each actuator is uniformly sampled between 16 to 20 $N \cdot m$. The magnitude of the random force is uniformly sampled between 0 to 10 $N$ and the direction is uniformly sampled from a spherical surface. The force is periodically applied and last for 0.2s each time.

\begin{figure} [t]
    \centering
    \includegraphics[width=0.8\linewidth]{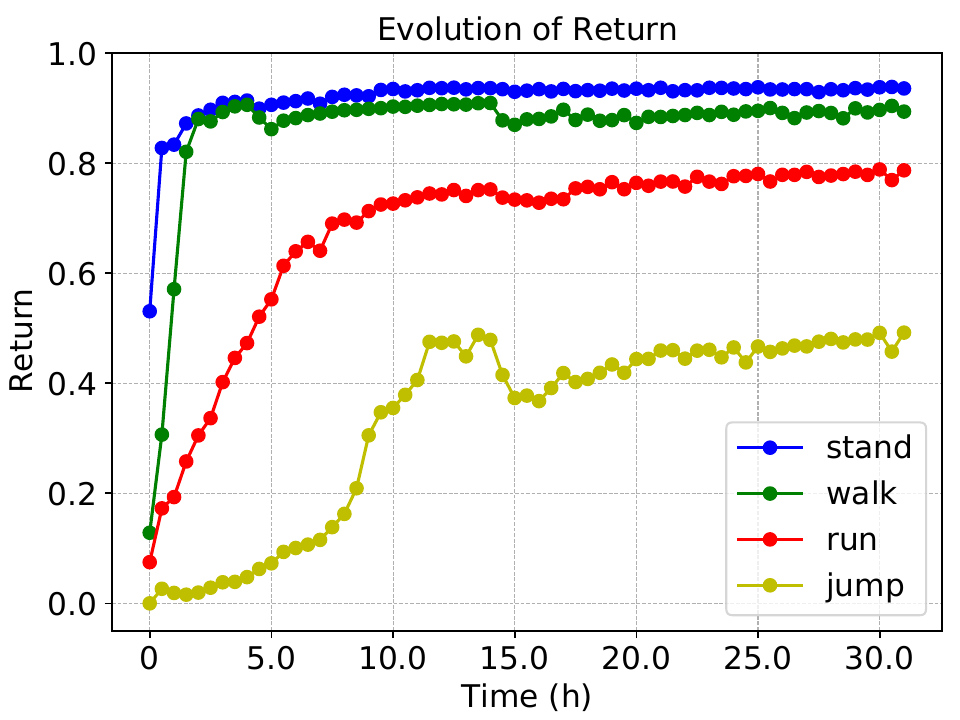}
    \caption{The evolution of imitation performance. Horizontal axis represents training time in hours, vertical axis represents average return. Colors represent different gaits.}
    \label{fig:figure_evolution}
    \vspace{-0.5cm}
\end{figure}

\section{EXPERIMENTAL RESULTS}

We use a quadrupedal platform independently designed by our team~\cite{chi2022linearization}. The quadruped consists of twelve actuators with three (hip, upper leg and knee) on each leg. Pybullet is used as the simulation environment during training~\cite{coumans2016pybullet}. A distributed reinforcement learning infrastructure named TLeague~\cite{sun2020tleague} is employed to increase training efficacy. We first show the results by enabling all the techniques we proposed and then compare the full method with alternatives that disable some components. We find that prioritized sampling and early termination are helpful for tracking the jumping trajectories and action space with change of joint angular velocity help the system to achieve robust tracking of the sprinting trajectories.

\begin{figure} [t]
    \centering
    \includegraphics[width=\linewidth]{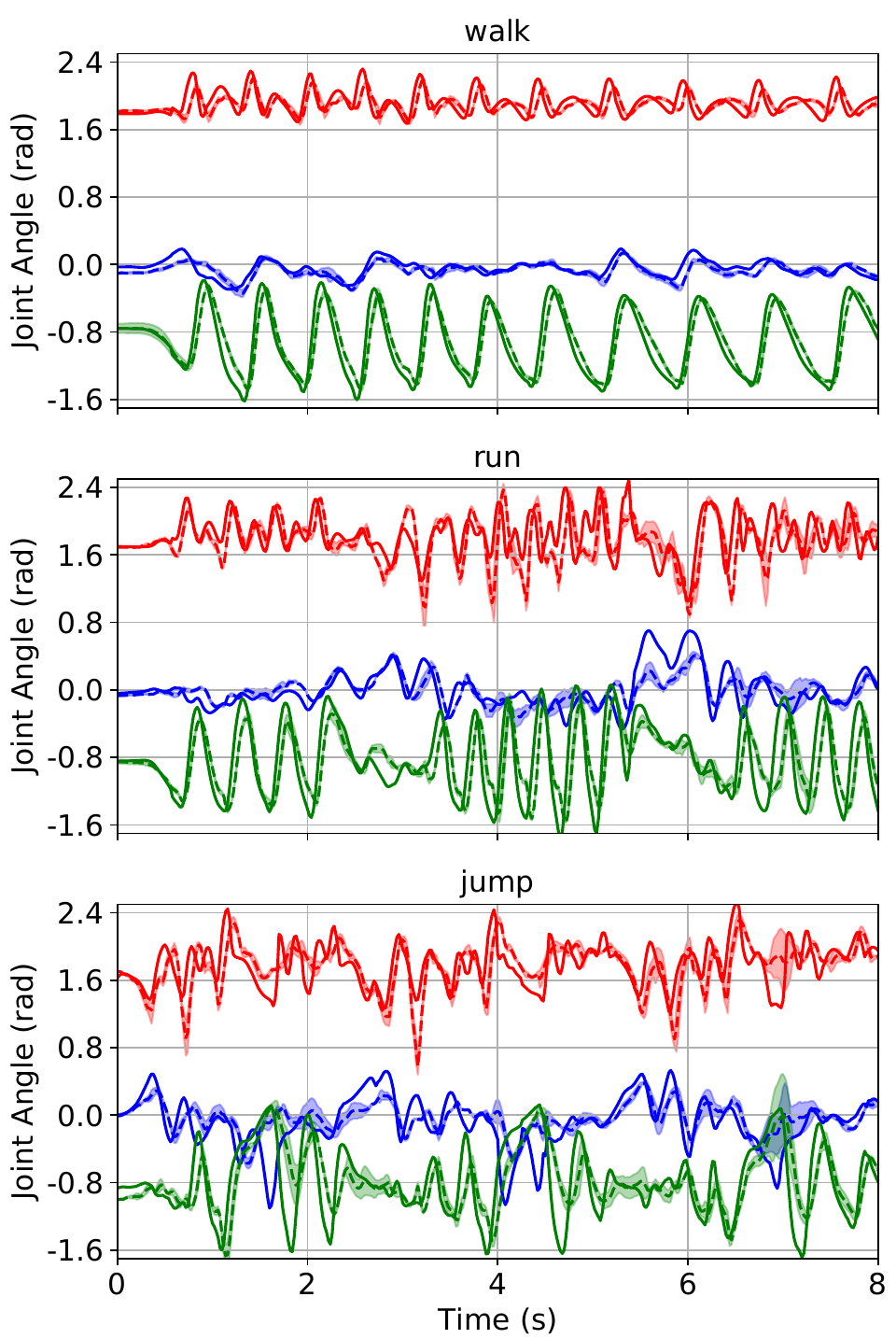}
    \caption{Joint angle curves tested at the end of training. Only curves from front left leg are shown for clarity. Top, mid and bottom panel represent curves of representative walking, running and jumping motion. Horizontal axis represents time in second and vertical axis represent joint angle in radian. Solid lines represents reference data, dashed lines represent the joint angles of real robot. Shaded areas represents the standard deviation computed over three trials. Hip joint, upper leg joint and knee joint are colored with red, blue and green respectively.}
    \label{fig:figure_joint_angles}
    \vspace{-0.5cm}
\end{figure}

\subsection{Motion Imitation}

The snapshots of learned behaviors are shown in Fig.~\ref{fig:figure_animation}. Simulation results are shown on top of real robot behaviors. In the simulation, the white robot is the kinematic reference data and the black robot is the physics-based animation. We can see that the robot is able to accurately imitate the reference data most of the time except during the falling stage of the jumping action.

\begin{figure} [t]
    \centering
    \includegraphics[width=0.8\linewidth]{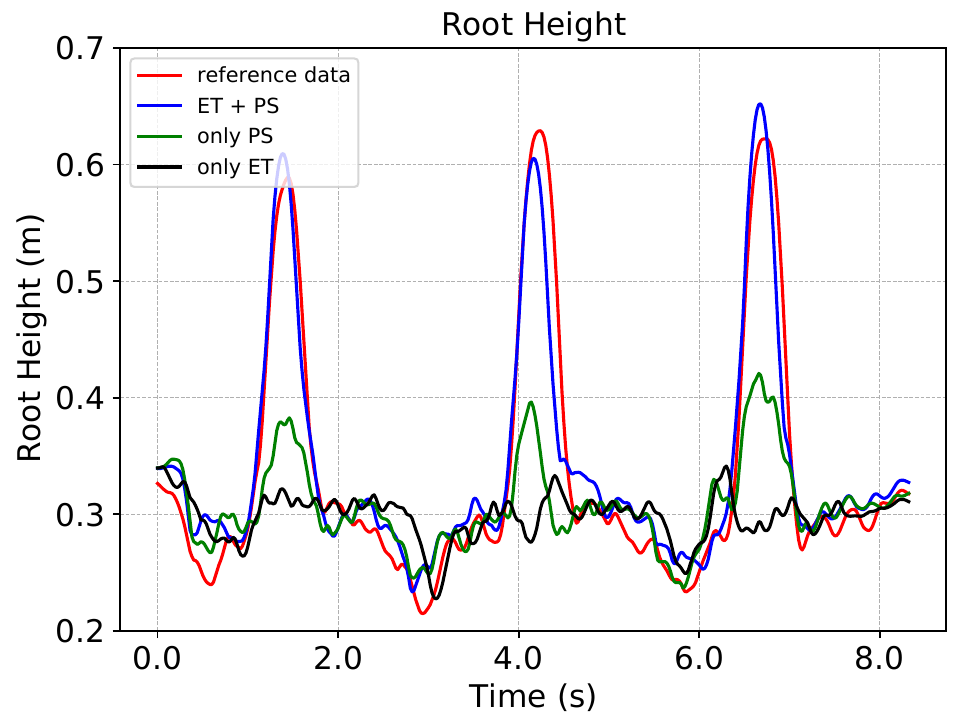}
    \caption{Jumping performance measured at the end of training. One representative motion data is used. The data lasts around eight seconds and includes three jumping actions. The performance is measured by the root height. Red curve represents reference data, blue curve is obtained by enabling prioritized sampling (PS) and early termination (ET), green and black curves are obtained by enabling either PS or ET.}
    \label{fig:figure_base_height}
\end{figure}

\begin{figure} [t]
    \centering
    \includegraphics[width=\linewidth]{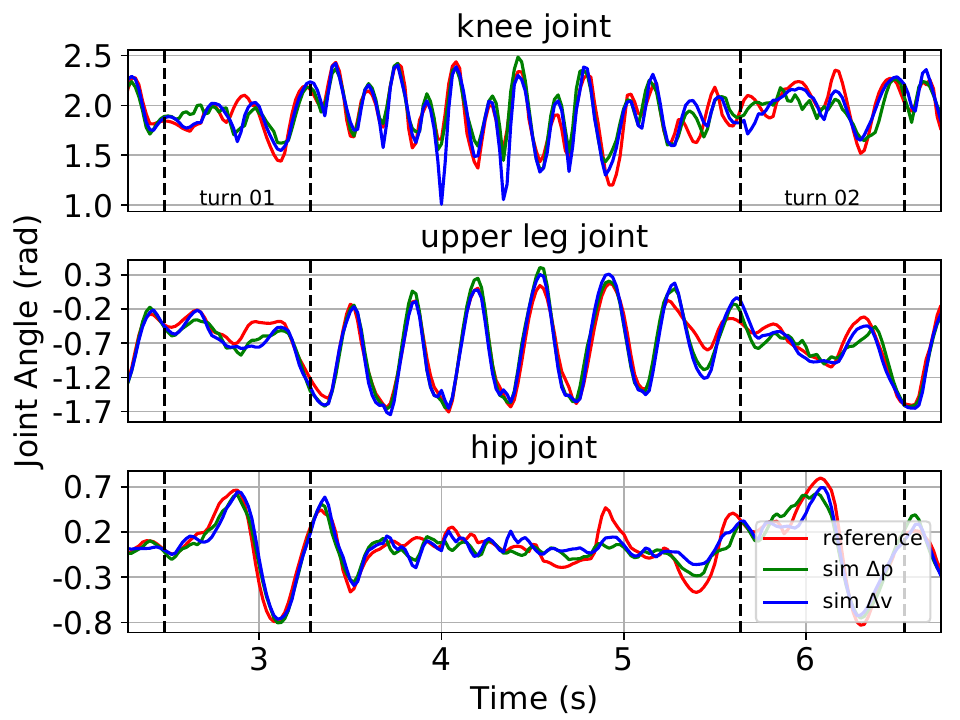}
    \caption{Joint angle curves of sprinting motion tested under different action space representations in simulation. Only curves of front left leg are shown for clarity. Knee joint, upper leg joint and hip joint are shown from top to bottom. Red curves are reference motion, green curves are obtained by action space representation with change of joint angle, denoted as $\Delta \text{p}$, blue curves are obtained by action space representation with change of angular velocity, denoted as $\Delta \text{v}$.}
    \label{fig:action_space}
    \vspace{-0.5cm}
\end{figure}

Fig.~\ref{fig:figure_evolution} shows the evolution of imitation performance obtained in simulation. During training, model parameters are saved at checkpoints and the performance is measured by the average reward over episodes and timesteps. Horizontal axis represents time in hours and vertical axis represents reward tested at corresponding checkouts. Different colors correspond to different gaits. We can see that standing and walking learn faster compared to high dynamic motion (running and jumping). The final performance of jumping is relatively low, it may be due to the physical constraints of the robot, such as the enforcement of motor limit and the misalignment of the body configuration between the quadrupedal robot and the real animal.

\begin{figure} [t]
    \centering
    \includegraphics[width=0.8\linewidth]{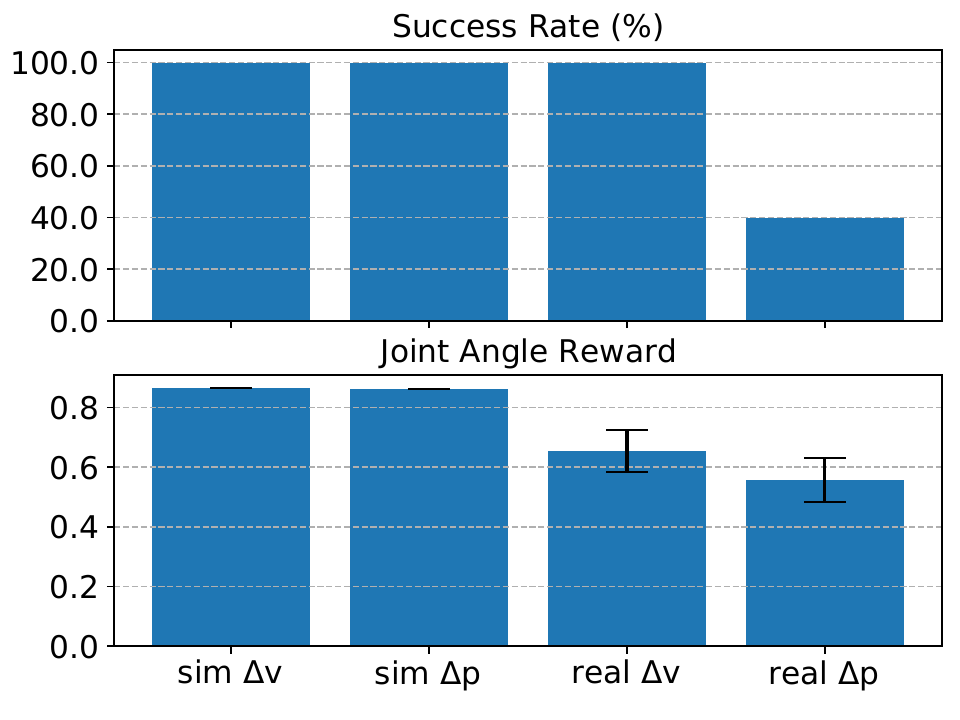}
    \caption{Performance of imitating sprint data under different conditions. The results are obtained both in simulation and real robot platform. Top panel shows the success rate and bottom panel shows the joint angle tracking accuracy. The arrow bars in bottom panel from real robot experiments are obtained from five trials.}
    \label{fig:figure_sprint}
    \vspace{-0.5cm}
\end{figure}

Fig.~\ref{fig:figure_joint_angles} shows the joint angle curves at the end of training tested on the real robot. We test the performance of walking, sprinting and jumping and show representative curves of each gait. Each panel in the figure represents the curves from front left leg, solid lines is the motion of reference data and dashed lines are from joint encoders of the real robot. Shaded areas are the standard deviations averaged over three trials. The curves of the other legs are similar with a slight difference between front and back legs. Hip joint, upper leg joint and knee joint are colored with red, blue and green respectively. Tracking of the joint angle is relatively accurate for low speed motion except with a phase shift. The phase shift may due to the communication delay of the real robotic system. As the speed of the motion becomes higher, the tracking performance gets worse, the trend is also reflected in the learning curve in Fig.~\ref{fig:figure_evolution}. Factors that may impact the performance include the discrepancy of the actuator model between the simulation and real robot and the simplified contact model used in simulation.


To evaluate the impact of prioritized sampling (PS) and early termination (ET) on the learned highly dynamic behaviors, we compare the full method with alternatives that disable some components. Fig.~\ref{fig:figure_base_height} plots the jumping performance of one representative motion data at the end of training tested in simulation. The tested data lasts around eight seconds and consists of three jumping actions. The performance is measured by the root height. The red curve represents the root height of reference data. The blue curve is probed from the model learned by enabling PS and ET. The green curve and black curves are obtained by enabling either PS or ET. We can see that enabling PS and ET allows the model to achieve consistent peak height with the reference data.

Fig.~\ref{fig:action_space} plots joint angle curves of sprinting motion tested under different action space representations in simulation. We select a representative sprinting data which consists of three straight runs and two sharp turnings. The peak base linear speed is around 4 m/s. Only curves of front left leg are shown for clarity. Knee joint, upper leg joint and hip joint are shown from top to bottom. Red curves are reference motion, green curves are obtained by action space representation with change of joint angle, denoted as $\Delta \text{p}$, blue curves are obtained by action space representation with change of joint angular velocity, denoted as $\Delta \text{v}$. The turning periods are shown between two black vertical dashed lines. We notice oscillations of front leg during turning motion when first trying $\Delta \text{p}$ representation, as reflected from the green knee joint curve. This representation is commonly used in previous work~\cite{peng2018deepmimic,tan2018sim} under low-speed motion and always accompanied with additional joint smooth term added to reward. The robot exhibits smoother motion when using $\Delta \text{v}$ representation, as reflected from the blue joint curves.

Fig.~\ref{fig:figure_sprint} compares the performance of imitating sprinting data under two action space representations. The results are obtained both in simulation and real robot platform. Top panel shows the success rate and bottom panel plots the joint angle tracking accuracy. We can see that in simulation, the two action space representation exhibit similar performance, while in real robot platform, the action space with change of joint angular velocity is superior than action space with change of joint angle. We repeat the experiment of tracking sprinting data five times and compute the success rate and joint angle reward. A successful trail is defined as completing the tracking without any falling. The most challenging period is the speed up phase after the sharp turning, the failure occurs during this period most of the time. The action space with change of joint angular velocity achieves robust tracking with higher accuracy.

\begin{figure} [t]
    \centering
    \includegraphics[width=\linewidth]{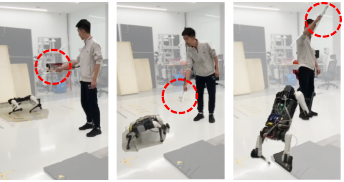}
    \caption{Snapshots of gait activation. Left: walk, middle: sit, right: jump. The stick with markers are labeled within the red dashed circle.}
    \label{fig:figure_interaction}
    \vspace{-0.5cm}
\end{figure}

\subsection{Gait Activation}

Fig.~\ref{fig:figure_interaction} shows snapshots of gait activation deployed on real robot. Humans can interact with the robot using a stick with markers attached to it. The robot follows the velocity of the stick in horizontal plane and the position of the stick in vertical axis and exhibits various behaviors. The motion of the stick is perceived using a motion capture system. Walking, sitting and jumping are shown from left to right. The stick with markers are labeled within the red dashed circle. The full video can be seen in the supplementary materials.

\begin{figure} [t]
    \centering
    \includegraphics[width=0.7\linewidth]{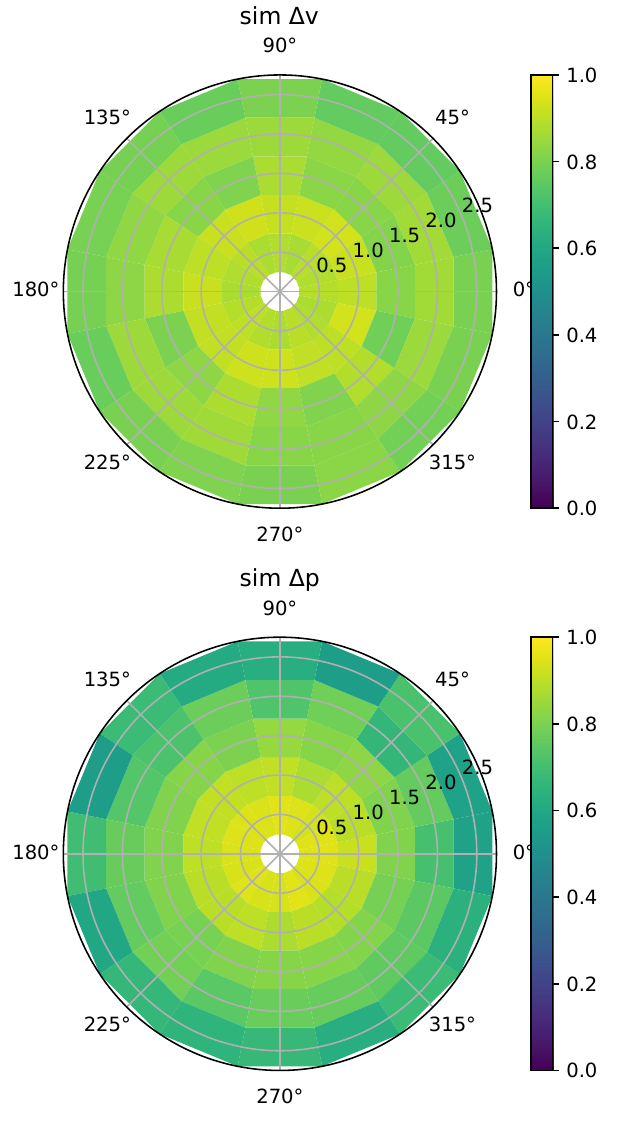}
    \caption{Polar plot of velocity following performance. Angle represents target direction, radius represents target speed and density represents the tracking reward. Top panel is the results obtained using action space with change of joint angular velocity, bottom panel is obtained using action space with change of joint angle.}
    \label{fig:figure_mid_polar}
    \vspace{-0.5cm}
\end{figure}

We compare the velocity tracking performance under two action space representations in simulation and Fig.~\ref{fig:figure_mid_polar} shows the results. We vary target direction and target speed and compute the velocity tracking reward as in~\cite{ling2020character}. Target speed is varied within range 0.5-3m/s with spacing 0.5m/s. Target direction in horizontal plane is varied within range 0-360 degrees with spacing 22.5 degrees. The initial orientation of the robot is aligned with the target direction. Fig.~\ref{fig:figure_mid_polar} is shown in polar coordinate, angle represents target direction, radius represents target speed and density represents the tracking reward. Top panel is the results obtained using action space with change of joint angular velocity, bottom panel is obtained using action space with change of joint angle. We can see that the overall performance of the top panel is better than bottom panel. In the bottom panel, as the target speed increase, the velocity tracking reward decrease, while in the top panel the performance is relatively robust to target speed changes.

\section{CONCLUSIONS}

In this paper, we propose a learning-based approach to enable robot to perform various highly dynamic skills and these skills can be activated through human interaction much the same way humans interact with a pet. The proposed approach can be robustly deployed on the real quadrupedal robot. One future direction could consider active compliance control when the robot performs highly dynamic behaviors. This would help to minimize the contact force induced by landing after jump motion or speeding up followed by sharp turning. Another possible extension of this work could be to enable vision-based perception, which would allow humans to interact with the robot with more flexibility.





\section*{APPENDIX}

The trajectory encoder and context encoder are both a multilayer perception (MLP) with 2 hidden layers, each has 256 hidden units and ReLU activation function. The output is a linear layer and has dimension of 32, which corresponds to the dimension of the embedding space. The embedding space is a categorical discrete space with dictionary size of \textit{K}=256. The decoder is also a MLP with 2 hidden layers, each has 256 hidden units and ReLU activation function. The output is a linear layer with dimension of 12, which corresponds to the number of actuators of the robot. The discriminator used in equation (\ref{eqn:context reward}) is a MLP with 2 hidden layers and 256 hidden units each layer. The weighting coefficients in the rewards are set as follows: $\alpha$=1, $\beta$=0.25, $\gamma$=1.


\newpage




\bibliographystyle{IEEEtran}
\bibliography{IEEEabrv}

\end{document}